\documentclass[lettersize,journal]{IEEEtran}
\usepackage{amsmath,amsfonts}
\usepackage{algorithmic}
\usepackage{algorithm}
\usepackage{array}
\usepackage[caption=false,font=normalsize,labelfont=sf,textfont=sf]{subfig}
\usepackage{textcomp}
\usepackage{stfloats}
\usepackage{url}
\usepackage{verbatim}
\usepackage{graphicx}
\usepackage{cite}
\begin{document}

\title{Autonomous Decision Making for UAV \\ Cooperative Pursuit-Evasion Game \\ with Reinforcement Learning}

\author{\IEEEauthorblockN{Yang Zhao, Zidong Nie, Kangsheng Dong, Qinghua Huang, Xuelong Li}}



\maketitle

\begin{abstract}
The application of intelligent decision-making in unmanned aerial vehicle (UAV) is increasing, and with the development of UAV 1v1 pursuit-evasion game, multi-UAV cooperative game has emerged as a new challenge. This paper proposes a deep reinforcement learning-based model for decision-making in multi-role UAV cooperative pursuit-evasion game, to address the challenge of enabling UAV to autonomously make decisions in complex game environments. In order to enhance the training efficiency of the reinforcement learning algorithm in UAV pursuit-evasion game environment that has high-dimensional state-action space, this paper proposes multi-environment asynchronous double deep Q-network with priority experience replay algorithm to effectively train the UAV's game policy. Furthermore, aiming to improve cooperation ability and task completion efficiency, as well as minimize the cost of UAVs in the pursuit-evasion game, this paper focuses on the allocation of roles and targets within multi-UAV environment. The cooperative game decision model with varying numbers of UAVs are obtained by assigning diverse tasks and roles to the UAVs in different scenarios. The simulation results demonstrate that the proposed method enables autonomous decision-making of the UAVs in pursuit-evasion game scenarios and exhibits significant capabilities in cooperation.
\end{abstract}

\begin{IEEEkeywords}
unmanned aerial vehicle, pursuit-evasion game, deep reinforcement learning, multi-role cooperation, autonomous decision-making.
\end{IEEEkeywords}

\section{Introduction}
\IEEEPARstart{T}{he} capability of unmanned aerial vehicle (UAV) continues to enhance through the utilization of advanced flight control, payload, power, and other technologies. This progress serves as a new driving force for its technological development and facilitates the rapid generation of UAV's game capability in highly challenging environments. The development of UAV equipment technology exhibits characteristics such as networking, decentralization, cost-effectiveness, and intelligence. With advancements in sensor technology, airborne computing power, and communication capabilities of weapons and equipment, the performance of UAV will witness further enhancements. Consequently, these low-cost and mass-produced UAV will find wider application across various scenarios. Equipped with autonomous decision-making capabilities, UAV can significantly contribute to areas including reconnaissance missions, manned-unmanned cooperation, as well as pursuit-evasion game.

At present, the research on UAV pursuit-evasion game primarily concentrates on 1v1 UAV game and multi-UAV cooperative game. In the field of 1v1 UAV pursuit-evasion game, there are three traditional method: game theory for modeling and solving pursuit-evasion game scenarios\cite{ref1,ref2,ref3,ref4}, optimization theory to model pursuit-evasion game as a multi-objective decision optimization problem\cite{ref3,ref4}, and utilizing artificial intelligence decision technology with self-learning capabilities\cite{ref5}. The game theory-based approach is limited by its myopic focus on short-term advantages in UAV game, and the difficulty of accurately modeling complex pursuit-evasion game scenarios. The computational performance of the pursuit-evasion game decision method, based on optimization theory, often fails to satisfy the real-time requirements of pursuit-evasion game decision-making and is primarily employed for offline research aimed at optimizing pursuit-evasion game policies. While the pursuit-evasion game situation exhibits significant diversity and the artificially generated rules are incapable of encompassing all conceivable scenarios. Consequently, while the method may appear straightforward, it necessitates a substantial workload and falls short in terms of both robustness and accuracy. The emergence of deep learning technology has led to significant advancements in various domains\cite{ref6, ref7, ref8, ref9}.  Reinforcement learning (RL), an artificial intelligence technique for intelligent decision-making, has merged with deep learning. In recent years, deep reinforcement learning has emerged as one of the most successful methodologies in the field of artificial intelligence, with widespread applications in intelligent decision-making, control and so on\cite{ref10, ref11, ref12, ref13, ref14, ref15}, and also plays a crucial role in intelligent pursuit-evasion game. By establishing a decision-making framework to govern the agent-environment interaction and formulating a rational reward function, deep reinforcement learning empowers the UAV agent to effectively acquire knowledge and make informed decisions in pursuit-evasion game scenarios. This not only enhances confront effectiveness but also bolsters survival capabilities, thereby attracting considerable attention from researchers in the field of intelligent pursuit-evasion game. In order to enhance the efficiency of reinforcement learning algorithms in exploring the policy space, Zhang et al.\cite{ref16} proposed a heuristic Q-network approach that incorporates expert knowledge to guide the search process. This method utilizes the heuristic Q-network technique to train neural network models for solving the over-the-horizon pursuit-evasion game maneuver decision problem. Yang et al.\cite{ref17} Proposed a approach that presents a decision-making method for autonomous maneuvers of UAV in pursuit-evasion game, utilizing the DDPG algorithm. This method effectively filters out numerous invalid action values using optimized pursuit-evasion game maneuver action values generated by the optimization algorithm. Furthermore, it incorporates the optimized action into the replay buffer as an initial sample, thereby enhancing both game effectiveness and survivability of the DDPG algorithm during UAV pursuit-evasion game.

The increasing complexity of the UAV application environment and the growing diversity of tasks have posed challenges for a single UAV to effectively handle various application scenarios. With the development of deep reinforcement learning technology in the multi-agent field\cite{ref18, ref19, ref20}, the cooperative technology of multiple UAVs has emerged as an imperative solution and a significant developmental trend. Based on the 1v1 UAV game, researchers have devoted their efforts to studying multi-UAV cooperative pursuit-evasion game. The decision-making problem of cooperative multi-target attack in pursuit-evasion game was investigated by Luo et al.\cite{ref21}, who proposed a heuristic adaptive genetic algorithm to effectively explore the optimal solution for missile target assignment. The proposed approach by Wang et al.\cite{ref22} employed the clonal selection algorithm to establish a multi-step UAV dynamic weapon-target assignment game model, based on the double matrix game Nash equilibrium point solution method, resulting in a more precise Nash equilibrium solution. Furthermore, the technology of deep reinforcement learning also finds extensive applications in the domain of multi-agent systems. Zhang et al.\cite{ref23} successfully implemented communication between UAVs through bidirectional recurrent neural networks, integrating target allocation and  pursuit-evasion game situation assessment to generate cooperative tactical maneuver strategies that merge formation tactical objectives with each UAV's reinforcement learning objective. Li et al.\cite{ref24} proposed a multi-agent double soft actor-critic algorithm, which employs a distributed execution framework based on decentralized partially observable Markov decision process and centralized training. It considers the multi-UAV cooperative pursuit-evasion game problem as a complete cooperative game in order to achieve effective collaboration among multiple UAVs. The aforementioned methods consider the communication and collaboration among multiple UAVs to effectively accomplish cooperative pursuit-evasion game missions. However, they regard the UAV formation as a whole, with only cooperation rather than detailed division of labor, focusing on winning the game while overlooking the cost of UAV’s during such games. These approaches may lead to even if the UAV formation gain the eventual triumph, but individual UAVs may be exposed to potential encirclement, causing losses.

This paper proposes a deep reinforcement learning-based cooperative game method for multi-role formation of UAVs to effectively address this issue, wherein each UAV is assigned distinct roles in the pursuit-evasion game to optimize victory rate and minimize the cost of game. The main contributions of this paper are as follows:

i) The proposed algorithm, MEADDQN, enhances the efficiency of data collection for interactive training in reinforcement learning and improves sample efficiency through PER;

ii) We designed reward shaping for two different UAV roles and conducted training, enabling them to proficiently perform pursuit and bait tasks respectively.

iii) We established multi-role UAV cooperative pursuit-evasion game framework and validated its effectiveness in scenarios involving 2v1, 2v2, and 3v2, yielding favorable outcomes.

The subsequent sections of the paper are structured in the following manner. Section 2 presents the UAV dynamics model and offers a comprehensive exposition of the pursuit-evasion game system. Section 3 presents the maneuvering decision algorithms employed by the opposing sides. Section 4 presents the components involved in constructing reinforcement learning models. Section 5 presents the training and testing of the model, which are demonstrated through simulation analysis. And the paper concludes with Section 6, presenting a comprehensive summary encompassing the entirety of the study.

\section{Description of UAV Pursuit-Evasion \\ Game System}
\subsection{UAV Dynamics Model}
The UAV dynamics model serves as the fundamental basis for comprehending the pursuit-evasion environment of UAV. This study aims to investigate the intelligent decision-making capabilities of UAV in such environments. Consequently, when establishing the UAV model, it is abstracted as a particle model and employs a 3 degree of freedom (3-DOF) control mode\cite{ref25}.

\begin{figure}[!t]
	\centering
	\includegraphics[width=2.5in]{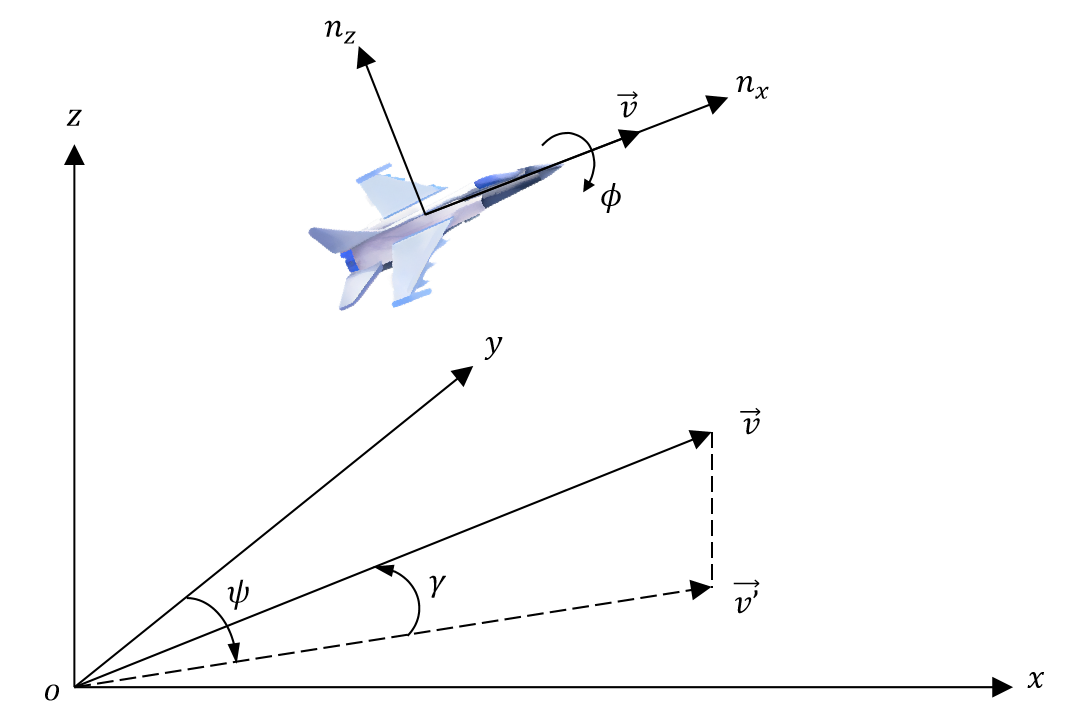}
	\caption{3-DOF UAV particle model.}
	\label{fig_1}
\end{figure}

In the inertial coordinate system, the state variables of the 3-DOF equation for UAV consist of $[x,y,z,v,\gamma,\psi]$, where $(x,y,z)$ represents the positional information of UAV in the inertial coordinate system, $v$ is a scalar denoting the velocity of UAV, $\gamma$ and $\psi$ are respectively indicative of the pitch angle and yaw angle of UAV, signifying its direction of motion. Where, $\gamma$ is defined as the angle between $\vec{v}$, the velocity vector of the UAV, and the x-o-y plane of the inertial coordinate system. $\psi$ is defined as the angle between $\vec{v'}$ and the y-axis, while $\vec{v'}$ is the projection of $\vec{v}$ onto the x-o-y plane of the inertial coordinate system.

The control variable of UAV can be represented by three parameters:$[n_{x} ,n_{z},\phi]$. Where, $n_{x}$ represents the overload in the direction of UAV velocity, which is used to control the acceleration and deceleration. The variable $n_{z}$ represents the vertical axis overload of the UAV body, while $\phi$ denotes the roll angle of the velocity vector, and they control the change of velocity direction collectively. The intelligent algorithm utilizes these three control variables to determine the maneuvering mode of the UAV, thereby enabling it to execute intricate aerial maneuvers and accomplish the pursuit-evasion game missions. Affected by these three control parameters, the changes in UAV’s speed, roll angle and yaw angle are as follows:
\begin{equation}\begin{cases}\dot{\nu}=g(n_x-\sin\gamma)\\\dot{\gamma}=\frac{g}{v}(n_z\cos\phi-\cos\gamma)\\\dot{\psi}=\frac{gn_z\sin\phi}{v\cos\gamma}\end{cases}\end{equation}

Furthermore, in the inertial coordinate system, the UAV coordinates exhibit the following variations:

\begin{equation}\begin{cases}\dot{x}=v\cos\gamma\sin\psi\\\dot{y}=v\cos\gamma\cos\psi\\\dot{z}=v\sin\gamma&\end{cases}\end{equation}

\subsection{Judgment Standard of Interception in Pursuit-Evasion Game}
The advantages and disadvantages of the confrontation in the pursuit-evasion game environment are conveyed via the relative situational information of the UAVs. It is expected that UAV can achieve intelligent decision-making to secure more advantageous firing positions during pursuit-evasion game. The coverage of a UAV's firepower typically forms a frontal cone, thereby enabling the determination of UAV's advantages and disadvantages in a pursuit-evasion game environment based on the UAV's orientation. The pursuit-evasion game environment discussed in this paper does not encompass the simulation of UAV firepower. Therefore, in order to effectively neutralize enemy aircraft, the UAV must autonomously maneuver and strategically position itself behind the target UAV, ensuring a tail chase to target is executed within UAV’s firepower range. This study imposes a strict numerical constraint on tracking interception in pursuit-evasion games.

\begin{figure}[!t]
	\centering
	\includegraphics[width=2.5in]{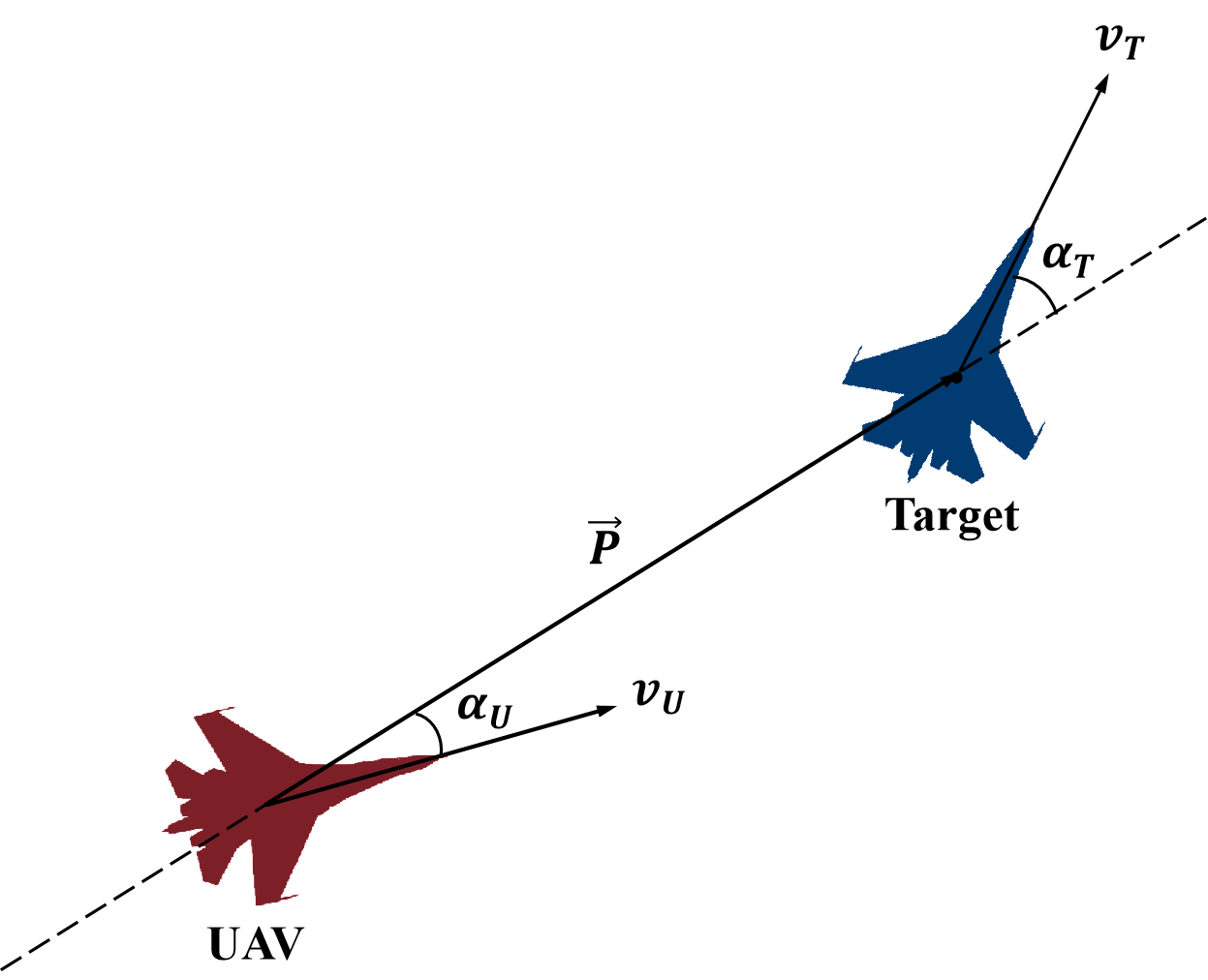}
	\caption{Judgment standard of Interception.}
	\label{fig_2}
\end{figure}

As shown in Fig. 2, the vector $\vec{P}$ represents the relative position of the UAV and the target, the antenna train angle $\alpha_{U}$ corresponds to the angle between the velocity vector $v_{U}$ of the UAV and the relative position vector $\vec{P}$, and the aspect angle $\alpha_{T}$ refers to the angle between the velocity vector $v_{T}$ of the target and the relative position vector $\vec{P}$. The UAV is judged to successfully intercept the target when $\alpha_U<5^\circ $, $\alpha_T<90^\circ $, and $d=\left\|\vec{P}\right\|<800m$.

\section{Maneuver Policy Modeling for UAV Pursuit-Evasion Game}
In the UAV pursuit-evasion game scenario described in this study, the UAV formations representing opposing factions are denoted as red and blue correspondingly. The red team policy network model in this paper is trained using the deep reinforcement learning algorithm to guide the red  UAVs in making maneuver decisions during cooperative pursuit-evasion game tasks within the red-blue formation. The blue UAVs realize their maneuver decision through the matrix game algorithm, serving as an adversary to evaluate the efficacy of the deep reinforcement learning algorithm.

\subsection{Red Team --- Multi-Environment Asynchronous  Double Deep Q-network Algorithm with Prioritized Experience Replay}

\begin{figure*}[!t]
	\centering
	\includegraphics[width=6.0in]{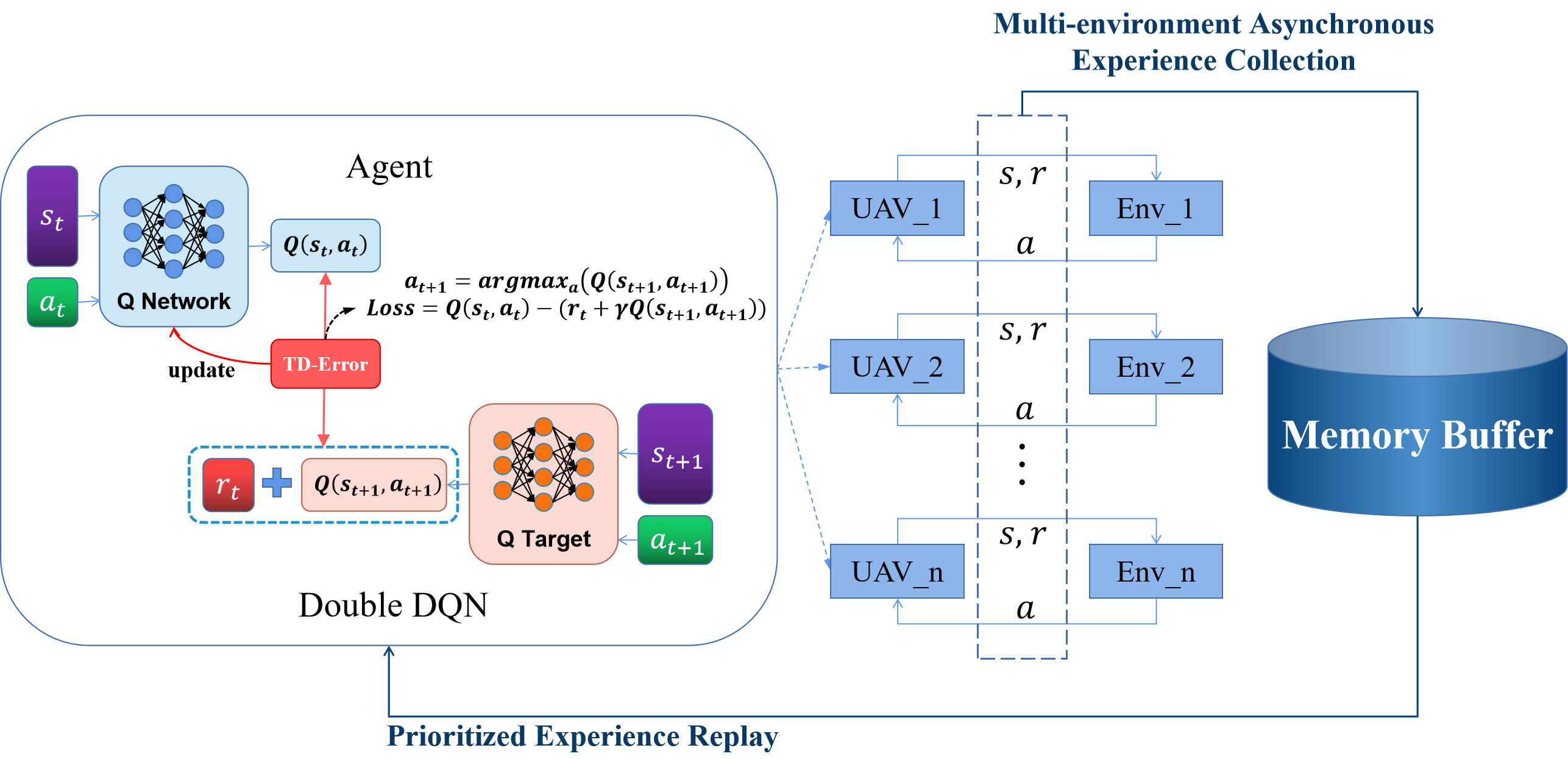}
	\caption{Reinforcement learning algorithm framework for UAV pursuit-evasion game.}
	\label{fig_3}
\end{figure*}

The field of reinforcement learning is dedicated to maximizing agent’s cumulative reward within a complex and uncertain environment. The agent improves its action selection by perceiving the environmental state and receiving rewarding feedback, thus obtaining the maximum return. Reinforcement learning problems are usually modeled utilizing Markov decision processes (MDP). MDP is a mathematical framework that models the decision-making process of an agent in an uncertain environment. It captures the notion that future states are determined solely by the current state and actions taken, without any dependence on past states. MDP can be represented using a quad-tuple:$\langle S,A,P,R\rangle $. The state space is denoted as $S$, the action space as $A$, the environment state transition probability as $P$, and the reward function $S\times A\to R$ quantifies the amount of feedback that agent can receive for executing an action in the current state.

\begin{equation}R(s,a)=\mathbb{E}\left[r_{t+1}\mid s_t=s,a_t=a\right]\end{equation}

The primary objective of reinforcement learning algorithms is to optimize strategies through interactive trial and error, with the ultimate goal of maximizing the returns.
\begin{equation}maximize\  G_t=\sum_{k=0}^{T-t}\gamma^kR_{t+k}\end{equation}
Where, $G_{t}$ is the returns, which is the cumulative discount reward after time t, and $\gamma$ is a discount factor that satisfies $0\leq\gamma\leq1$. The aforementioned definition can be intuitively comprehended as
an agent focusing more on near-term rewards than on rewards that are further away.

Reinforcement learning algorithm optimizes a policy $\pi$, $\pi{:}S\to A$ is the mapping function of the agent from state to action. Via maintaining an action value function $Q_{\pi}(s,a)$, the value-based reinforcement learning algorithms evaluate the benefit of selecting action $a$ in state $s$ when agent’s policy $\pi$ is determined.

\begin{equation}
	Q_\pi(s,a)=E_\pi[\sum_{k=0}^\infty\gamma^kR_{t+k+1}|S_t=s,A_t=a]
\end{equation}

The action-value function is updated iteratively according to the Bellman Equation (\ref{eq:6}), and $Q^*(s,a)=\max_\pi Q_\pi(s,a)$ is obtained through constantly approximation. Based on this, agent can obtain the optimal policy $\pi^*(a|s)=arg\max_{a\in\mathcal{A}(s)}Q^*(s,a)$.

\begin{equation}
	\label{eq:6}
	Q_\pi(s,a)=\mathbb{E}_\pi[R_{t+1}+\gamma Q_\pi(S_{t+1},A_{t+1})|S_t=s,A_t=a]
\end{equation}

Mnih et al. \cite{ref11} utilized $Q(s,a;\theta)$ function to approximate the optimal $Q^*(s,a)$ function and employed a deep neural network to solve for $Q(s,a;\theta)$, which forms the fundamental concept of the Deep Q-Network (DQN) algorithm\cite{ref26,ref27}. In order to improve the efficiency and stability of the algorithm, a target network $Q(s,a;\theta')$ is added, which participates in the training of the policy network and replicates the current parameters of the policy network at regular intervals. The target network in the training process introduces a certain time delay to decouple the value estimation of adjacent moments, thereby mitigating the impact of unstable fluctuations in data transmission during each iteration. In Double DQN \cite{ref28} algorithm, the loss function of the neural network is
\begin{equation}\mathcal{L}(\theta)=[Q(s_{t},a_{t};\theta)-r_{t}+\gamma Q(s_{t+1},a_{max};\theta^{\prime})]^{2}\end{equation}
The solution can be obtained by gradient descent method.

Double DQN is an off-policy reinforcement learning algorithm that can utilize a distinct policy for data acquisition, which differs from the current update policy. The interactive data will be stored in the replay buffer as the form of transition transition $(s_t,a_t,r_t,s_{t+1})$ and trained using the time series difference method, thereby effectively enhancing data utilization. Prioritized experience replay (PER) \cite{ref29} is a method for sampling interactive data. When storing each transition in the replay buffer, PER assigns different priorities to each transition based on the absolute value of its TD-Error $|\delta_t|$ (\ref{eq:9}), and selects the transition with higher priority for training with a higher probability during sampling.
\begin{equation}
	\label{eq:9}
	\delta_t=R_t+\gamma V(S_{t+1})-V(S_t)
\end{equation}

This section proposes a Multi-Environment Asynchronous Double Deep Q-Network (MEADDQN) algorithm, which serves as a further optimization of the aforementioned algorithm through the introduction of multi-environment asynchronous experience collection, with the objective of expediting the training process and enhancing the efficiency of acquiring pursuit-evasion game interaction data. As shown in Fig. 3, MEADDQN concurrently generates multiple pursuit-evasion game environments with identical tasks in parallel threads, each environment being initialized differently. Agents operate asynchronously and interact with distinct environments while adhering to the same policy network. All interaction data collected from these environments is consolidated into a unified replay buffer that supports PER, enabling sampling of data from the buffer during training. To enhance the algorithm's robustness and explore strategic possibilities, this paper introduces different action noise to the UAV agents in diverse interactive environments. In environments with higher levels of action noise, agents will engage in more audacious exploration, whereas in environments with lower levels of action noise, agents will leverage their acquired experience to identify the most rewarding decision within the current policy.

\subsection{Blue Team --- Matrix Game Algorithm}
This paper improves the UAV matrix game algorithm \cite{ref30} for multi-UAV. In the multi-UAV matrix game algorithm, the blue UAV constructs a $b\times r$-dimensional matrix $\mathcal{G}$ for each red UAV, where $b$ represents the number of available actions for the blue UAV and $r$ represents the number of available actions for the red UAV. The value of $\mathcal{G}_{ij}$ in this matrix represents the reward score for the blue side, based on the current pursuit-evasion game situation, assuming that the blue side take action $i$ and the red side take action $j$, following a similar approach as reinforcement learning's reward function to ensure fairness in subsequent adversarial games. After obtaining these payoff matrices, further processing is carried out:

i) Find the row minimum of each row of each payoff matrix and sum according to the row index $i$;

ii) Select the maximum value from these minimum value sums;

iii) The action $a_{i}$ corresponding to the maximum value's row index is the optimal maneuver of the blue side.

The row minimum represents the lowest return for blue UAV, assuming that red UAVs’ policy can always minimize blue UAV's return, while the maximum value of these minimum values ensures that blue UAV will receive the highest possible return even if red UAVs always choose actions unfavorable to blue UAV.

\section{Reinforcement Learning Element in \\ Multi-Role UAV Pursuit-Evasion Game}
The reinforcement learning problem is typically described using the MDP 4-tuple model $\langle S,A,P,R\rangle $, where the state transition probability P is determined by the environment itself and does not necessitate explicit modeling in model-free reinforcement learning algorithms. Therefore, this section will provide a detailed description of modeling the state space, action space, and reward function for the reinforcement learning task of multi-role UAVs cooperation.

\subsection{State Space}

The state space in this paper is designed to encompass all the state information of both UAVs, as well as variables that can express the relative information between the two opposing sides. This comprehensive representation serves as input for the policy network, enabling it to make informed decisions in the current confrontational scenario. UAV's state information can be characterized by its position, pitch angle, and yaw angle. Furthermore, the variables illustrated in Fig. 2 can also depict the relative information during pursuit-evasion game. The efficacy of reinforcement learning training is ensured by representing the state of UAV in this paper as a 13-dimensional variable: 
\begin{equation}(z_U,v_U,\gamma_U,\psi_U,z_T,v_T,\gamma_T,\psi_T,\alpha_U,\alpha_T,d,\gamma_P,\psi_P)\end{equation}

The first four quantities represent the attributes of the UAV itself: $z_{U}$ denotes the altitude, $v_{U}$ is a scalar that represents speed, $\gamma_{U}$ signifies the pitch angle, and $\psi_{U}$ indicates the yaw angle. The subsequent four quantities depict the characteristics of the target UAV: $z_{T}$ refers to its altitude, $v_{T}$ denotes its speed, $\gamma_{T}$ represents its pitch angle, and $\psi_{T}$ indicates its yaw angle. The remaining five quantities are employed to represent the relative information between the two drones, where $\alpha_{U}$, $\alpha_{T}$ and $d$ are introduced in Section II as indicators for antenna train angle, aspect angle and distance respectively. Additionally, as for the relative position vector $\vec{P}$, its numerical magnitude is indicated by $d$, and the orientation of $\vec{P}$ can be represented in a similar manner to the pitch and yaw angles of the UAV point model.

\subsection{Action Space}
The present study utilizes the body axis direction overload, longitudinal overload, and roll angle as control variables to establish a 3-DOF UAV particle control model. This model facilitates more accurate simulation of realistic flight trajectories. The present section introduces a 15-dimensional discrete action space specifically tailored for the DDQN algorithm in the context of reinforcement learning with discrete control\cite{ref31}. This customized discrete action space aims to accommodate the three control variables $(n_x,n_z,\phi)$, which is shown as Fig. 4.
\begin{figure}[!t]
	\centering
	\includegraphics[width=2.5in]{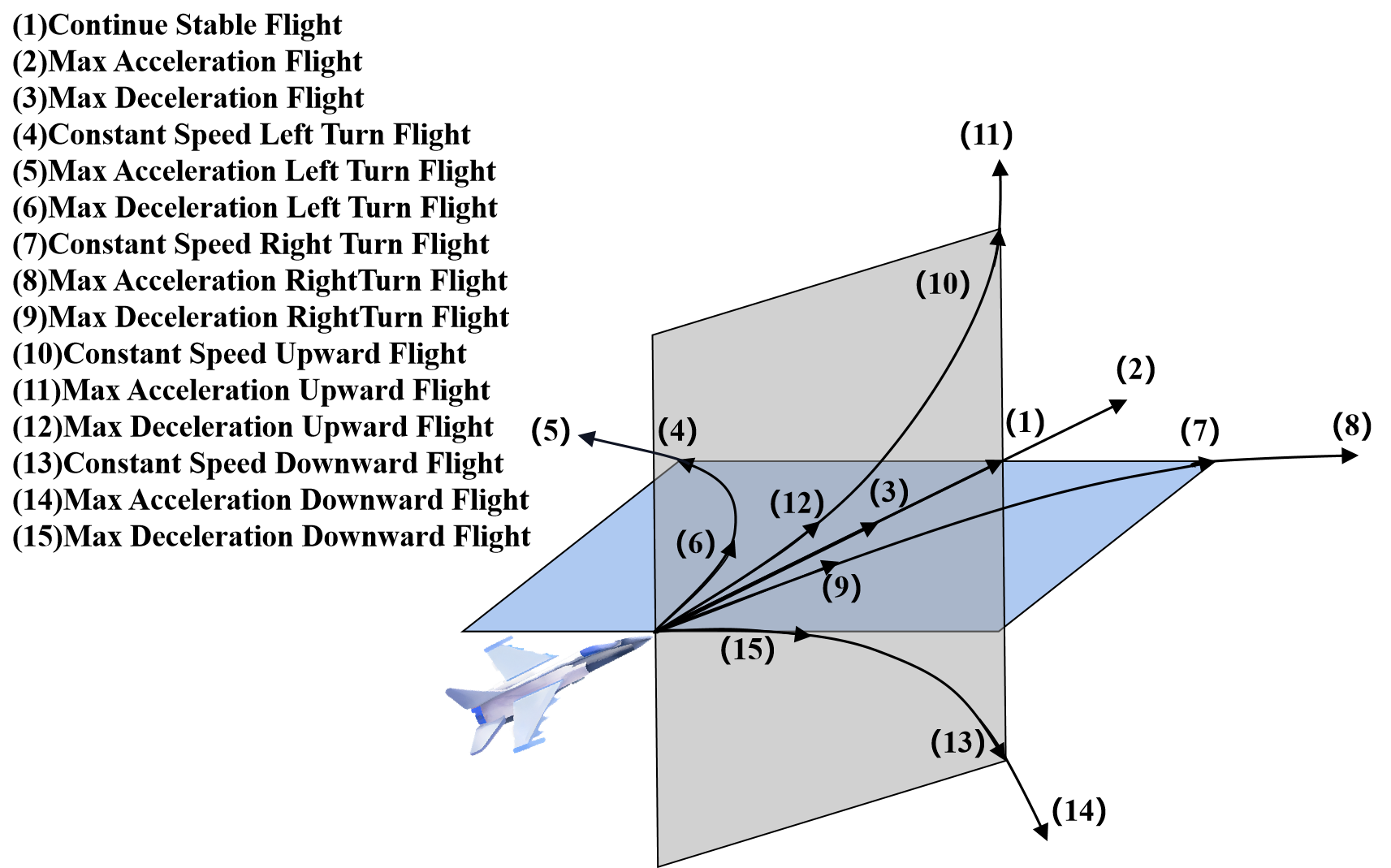}
	\caption{Autonomous decision action space for UAV.}
	\label{fig_4}
\end{figure}

\subsection{Reward Function for Multi-role UAV}
The reward function design in the task of UAV cooperative pursuit-evasion game aims to strategically guide the victory of the pursuit-evasion game. Thus the final outcome $r_{final}$ which signifies the victory or defeat of pursuit-evasion game, can be utilized directly as a reward signal. However, the agent is only provided with $r_{final}$ at the end of each episode, necessitating a prolonged waiting period to ascertain the correctness of its actions. Moreover, identifying advantageous action paths in environments with sparse reward poses a formidable challenge for the agent. This section enhances the efficiency of reinforcement learning algorithms through the utilization of dense reward shaping for the pursuit-evasion game task. The effective allocation of tasks is critical strategies for enhancing the win rate of pursuit-evasion game and minimizing operational losses. The UAV entities in the pursuit-evasion game environment are assigned the following roles: one type is designated for target attack and pursuit, called pursuit UAV, while the other type, called bait UAV, functions as a bait to draw enemy fire and create one-on-one or even multi-on-one scenarios for other UAVs. This section tailors the dense rewards in distinct manners for these two categories of UAVs, ensuring their ability to successfully accomplish their respective tasks.

\subsubsection{Pursuit UAV Reward Shaping}
\

$\textcircled{1}$Angle advantage reward:
\begin{equation}r_{p}=1-\frac{\alpha_{U}+\alpha_{T}}{2\pi}\end{equation}

The angle advantage reward aims for that $\alpha_{U}$ and $\alpha_{T}$ should be minimized, which aligns with the angle requirements of the judgment standard of interception.

$\textcircled{2}$Distance advantage reward:
\begin{equation}
	\label{eq:11}
	r_d=\exp\left(-\frac{abs(\|\vec{P}\|-d_{opt})}{d_0}\right)
\end{equation}

The distance advantage reward is designed to guide the UAV to reach the objective distance to the target. $d_{opt}$ in (\ref*{eq:11}) represents the objective distance, set to $d_{opt}=800m$, and $d_{0}$ is a distance constant parameter.

$\textcircled{3}$Velocity advantage reward:
\begin{equation}r_{v}=\frac{\overrightarrow{v_{U}}\cdot\vec{P}}{V_{max}\|\vec{P}\|}\end{equation}

The velocity advantage reward is directly proportional to the projection of the UAV’s velocity vector $v_{U}$ onto the relative position vector $\vec{P}$. The range of the velocity advantage reward is constrained to [-1,1] by $v_{max}$, which aligns with the range of the other two rewards.

To sum up, combining the collision and out-of-bound penalty term $r_{punish}$, the pursuit UAV's dense reward $r_{t}$ design is as follows:
\begin{equation}\begin{aligned}\\r_t=\begin{cases}r_{final} ,\quad&intercept \  or \ be \ intercepted\\r_{punish} ,\quad&collision \ or \ out \ of \ bound\\w_1r_p+w_2r_d+w_3r_v ,\quad&otherwise\end{cases}\end{aligned}\end{equation}
$w_{1}$, $w_{2}$, and $w_{3}$ are the weights of each reward.

\subsubsection{Bait UAV Reward Shaping}
\

$\textcircled{1}$Angle advantage reward:
\begin{equation}
	\label{eq:14}
	r_p=2*\exp\left(-\frac{abs(\alpha_T-\alpha_{opt})}{\alpha_0}\right)-1
\end{equation}

(\ref{eq:14}) represents that bait UAV needs to maintain an aspect angle to have a sufficient decoy effect on the target UAV, where $\alpha_{opt}$ is bait UAV’s objective aspect angle, and $\alpha_{0}$ is an angle constant parameter.

$\textcircled{2}$Distance advantage reward:
\begin{equation}r_d=\exp\left(-\frac{abs(\|\vec{P}\|-d_{opt})}{d_0}\right)\end{equation}

The distance advantage reward utilizes the identical calculation methodology as the pursuit UAV, wherein the objective distance is designated as $d_{opt}=1500m$. This ensures the safety of UAV while concurrently generating a decoy effect against the target.

The bait UAV does not necessarily require a consistent velocity advantage, but rather should be strategically positioned to allure the target. Therefore, considering the penalty term $r_{punish}$ for collision and out-of-bound situations, the dense reward $r_{t}$ for the bait UAV is designed as follows:
\begin{equation}r_t=\begin{cases}r_{final} ,&\quad be \ intercepted\\r_{punish} ,&\quad collision \ or \ out \ of  \ bound\\w_1r_p+w_2r_d ,&\quad otherwise\end{cases}\end{equation}

\section{Experiment}
\begin{figure}[!t]
	\centering
	\includegraphics[width=3.3in]{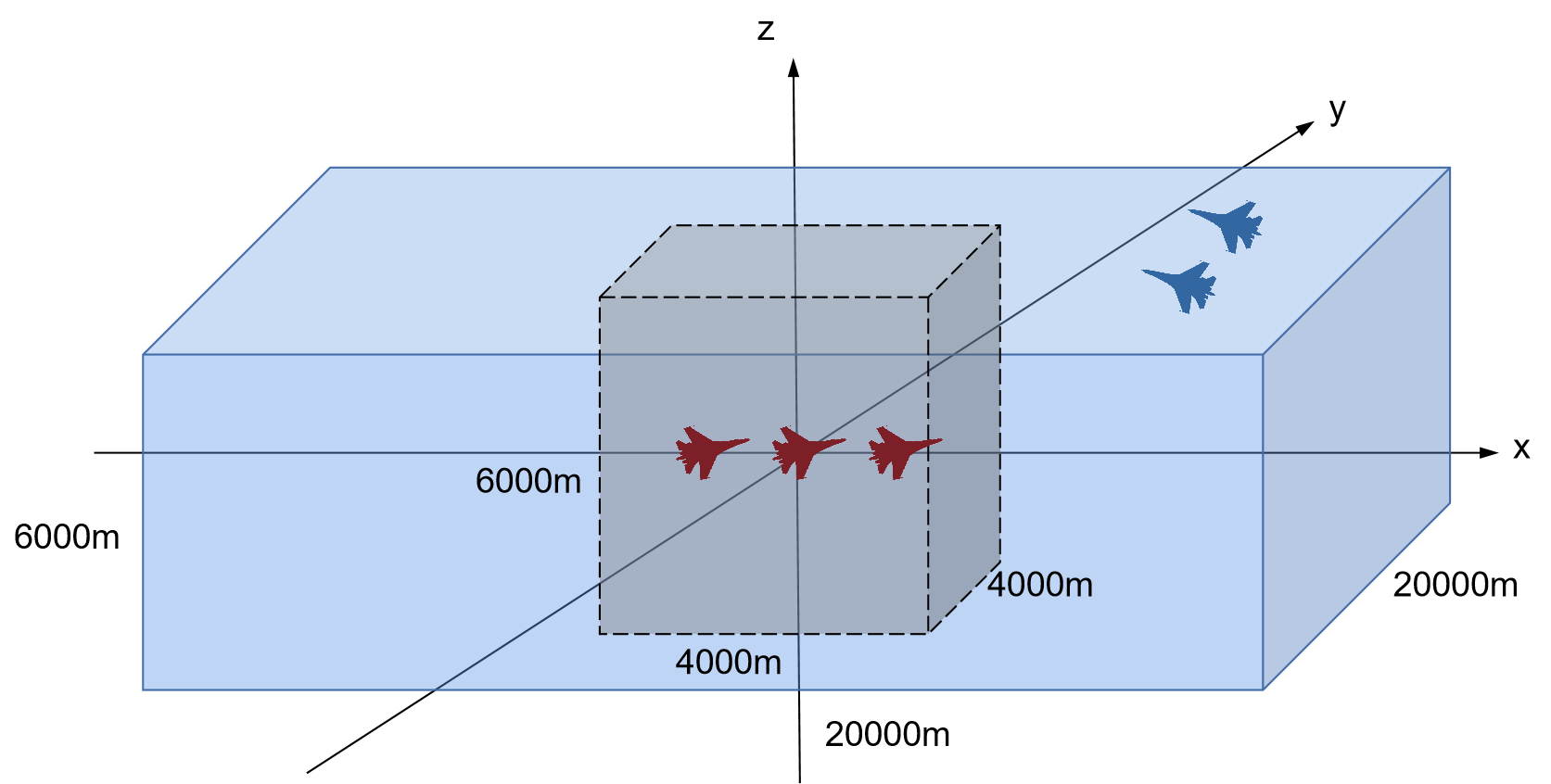}
	\caption{Initialization of environment situation.}
	\label{fig_5}
\end{figure}

In this study, pursuit UAV and bait UAV were trained independently and subsequently integrated within the multi-UAV environment. Both agent types use the same policy network architecture comprising three hidden layers with 512, 1024, and 512 neurons respectively, facilitating the policy transformation in a multi-UAV environment. In reinforcement learning training, the discount factor is $\gamma=0.95$, the replay buffer capacity is 100000, the batch size is 1024, and the activation function is ReLU. 

To faithfully replicate actual UAV flight conditions, no horizontal constraints are enforced within the airspace where pursuit-evasion game takes place while only the z-axis boundary is set as $1000m<z<13000m$ in the inertial coordinate system. In the designated airspace, a specified number of red and blue UAVs are deployed for pursuit-evasion game by their own maneuver policy.  The outcome is determined based on the cost of the game between the two sides' UAVs. Once a UAV is successfully intercepted by the opponent, we consider it destroyed and remove it from the pursuit-evasion game environment. The team that successfully intercepts all of the opponent first shall emerge as the victor. The initialization process of the pursuit-evasion game environment involves fixing the position of the red UAVs and establishing a spatial coordinate system with a red UAV as its origin. Then the blue UAVs should be initialized randomly within a rectangular space centered on the red UAV, which is limited 20000m in length, 20000m in width, and 6000m in height, as shown in Fig. 5, to ensure diversity in the initial position. To avoid the issue of decision steps becoming excessively short in subsequent rounds due to initial distances being too close, this study implemented an initial vacuum zone within a rectangular region measuring 4000 meters in length, 4000 meters in width, and 6000 meters in height, as shown in Fig. 5. The blue UAVs will not be initialized in this area, thereby ensuring a certain distance is maintained between both sides during the initial phase of the confrontation. As for the initialization of the flight attitude of UAVs, this study assumes that both UAVs start with a horizontal flight position. The red UAVs’ initial yaw angle is set to $\psi_{U}=0$, while the yaw angle of the blue team drone is randomly initialized. In this way, the simulated confrontation allows for various initial postures between the two side UAVs, thereby simulating states of advantage and disadvantage.

\subsection{Basic Training of Pursuit UAV and Bait UAV.}

\begin{figure}[!b]
	\centering
	\subfloat[]{\includegraphics[width=3.0in]{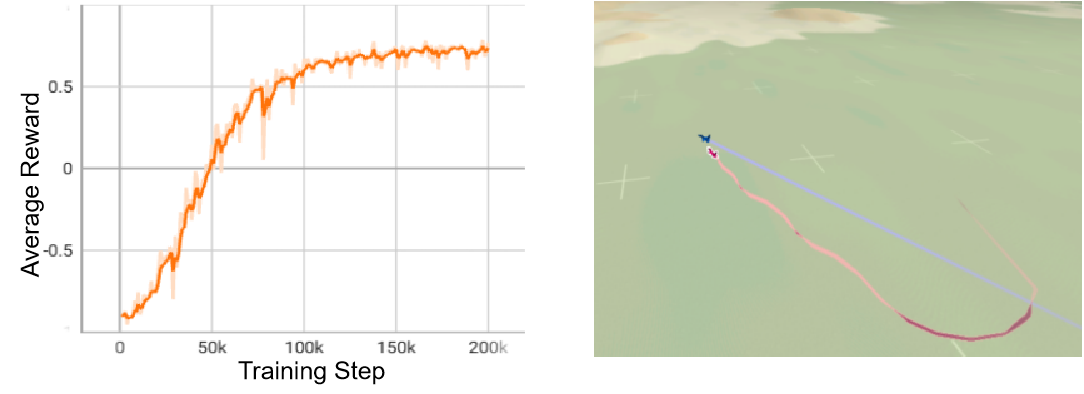}%
		\label{fig6_case1}}
	\hfil
	\subfloat[]{\includegraphics[width=3.0in]{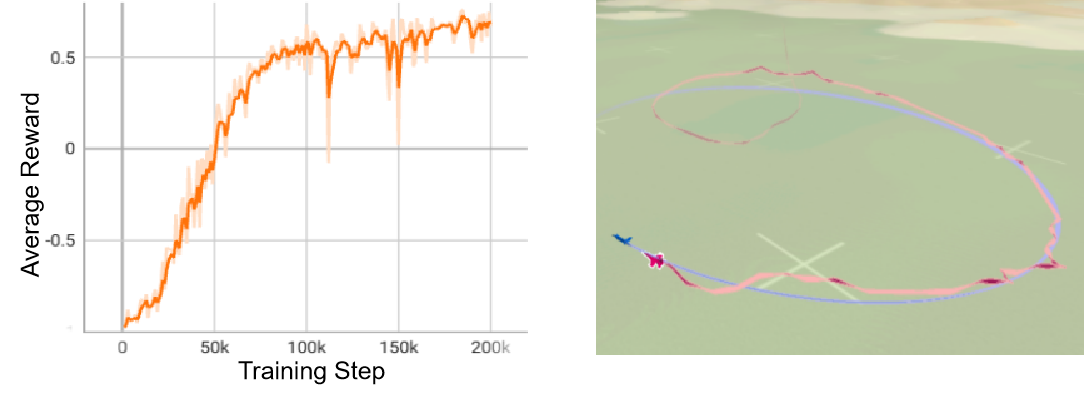}%
		\label{fig6_case2}}
	\hfil
	\subfloat[]{\includegraphics[width=3.0in]{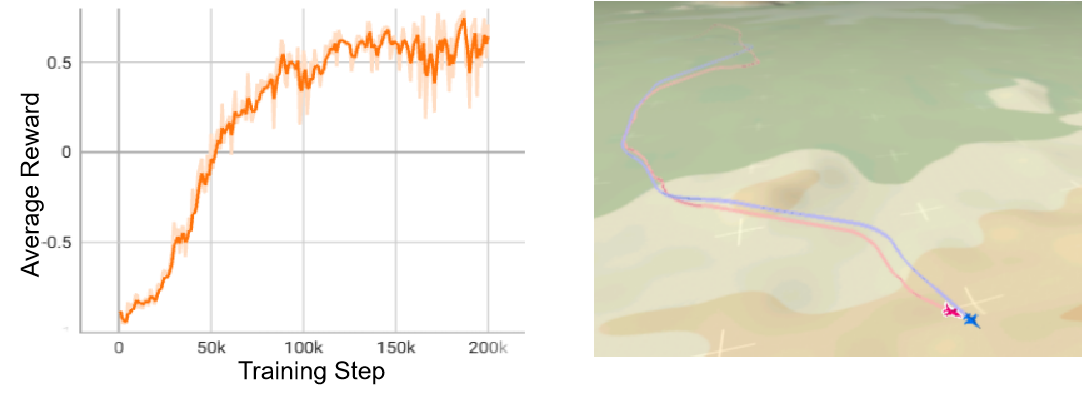}%
		\label{fig6_case3}}
	\caption{Pursuit UAV basic training reward curve and simulation. (a) against straight-line maneuver target. (b) against circling maneuver target. (c) against random maneuver target}
	\label{fig_6}
\end{figure}

\begin{figure}[!t]
	\centering
	\subfloat[]{\includegraphics[width=3.0in]{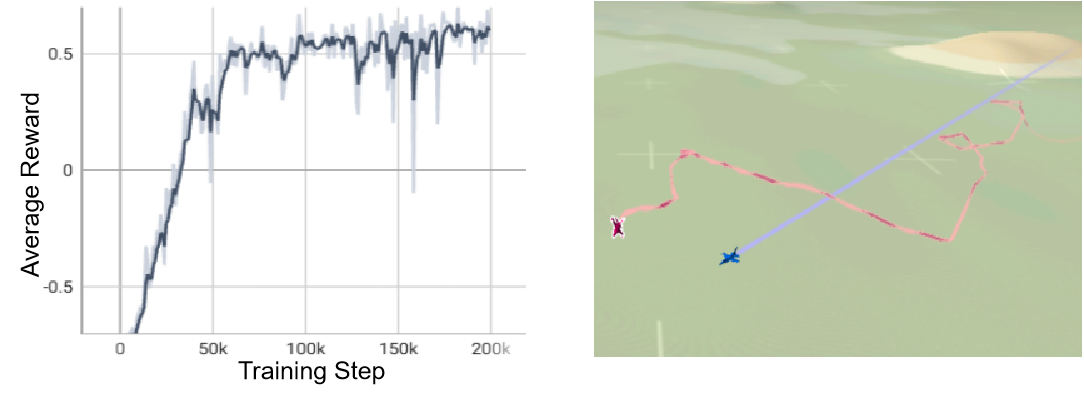}%
		\label{fig7_case1}}
	\hfil
	\subfloat[]{\includegraphics[width=3.0in]{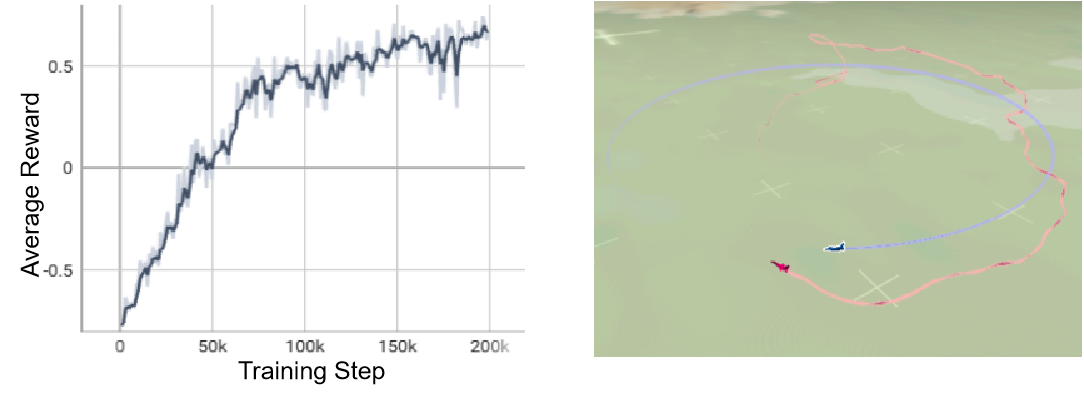}%
		\label{fig7_case2}}
	\hfil
	\subfloat[]{\includegraphics[width=3.0in]{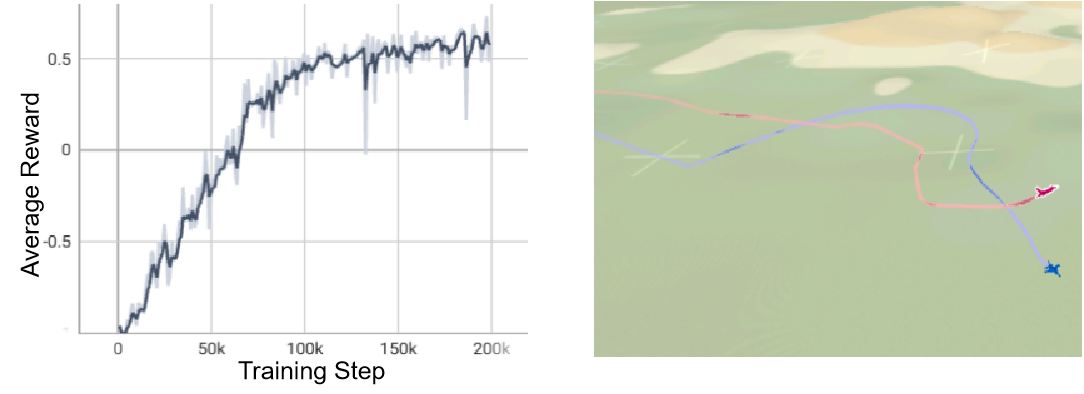}%
		\label{fig7_case3}}
	\caption{Bait UAV basic training reward curve and simulation. (a) against straight-line maneuver target. (b) against circling maneuver target. (c) against random maneuver target.}
	\label{fig_7}
\end{figure}

In the initial stages of training, the policy network of the red team is initialized randomly. However, this can lead to consistent failures for the red team and even difficulties in obtaining positive reward signals when directly confronted with a blue team that possesses a higher level of intelligent decision-making capability. Consequently, not only does this situation impact training efficiency but it also results in an exceedingly slow convergence speed. To solve this problem, this study proposes three basic training methods for the red team policy network. These basic training methods are intended to enable the red team's policy network to learn the maneuver model of the drone and the basic logic of pursuit-evasion game. Through a progressive sequence from simplicity to complexity, these three training methods enable the red team's policy network to gradually assimilate knowledge pertaining to UAV pursuit-evasion game. This not only enhances the intelligence level of the UAV agent but also establishes a solid groundwork for competing against highly intelligent decision-making adversaries. The three basic training scenarios are as follows: $\textcircled{1}$ against straight-line maneuver target; $\textcircled{2}$ against circling maneuver target; $\textcircled{3}$ against random maneuver target.

First, pursuit UAV’s policy network undergoes sequential basic training using MEADDQN with PER for three basic sessions, each consisting of 200000 steps. Finally, all these basic trainings are amalgamated, with one randomly initialized in each round for a total training duration of 600000 steps. Fig. 6 demonstrate the reward curves of three basic trainings and the pursuit-evasion game simulation after training completion respectively. It can be observed that with increasing complexity of the training scenario, there is a slight degradation in convergence. However, the pursuit UAV still effectively accomplishes its task.

Next, perform the same basic trainings for bait UAV, the results of which are shown in Fig. 7. Bait UAV is effectively maintaining a advantageous position ahead of the target, ensuring both safety and attractiveness.

\subsection{Against Matrix Game Algorithm Training.}

After successfully completing the three basic training sessions, it can be inferred that the red UAV’s policy network possesses a rudimentary comprehension of the 3DOF particle model of drones and the fundamental principles of pursuit-evasion game, and exhibits certain game capabilities. Consequently, it can game with blue UAV that employs matrix game algorithm and possesses intelligent decision-making ability. Fig. 8 shows the training reward curves of the pursuit UAV and bait UAV in a pursuit-evasion game against UAV controlled by matrix game algorithm. 

\begin{figure}[!t]
	\centering
	\subfloat[]{\includegraphics[width=1.65in]{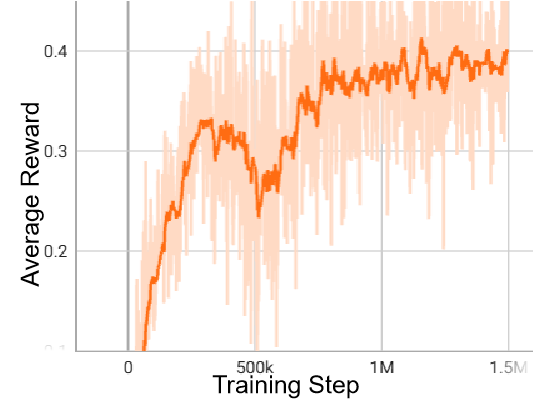}%
		\label{fig8_case1}}
	\hfil
	\subfloat[]{\includegraphics[width=1.65in]{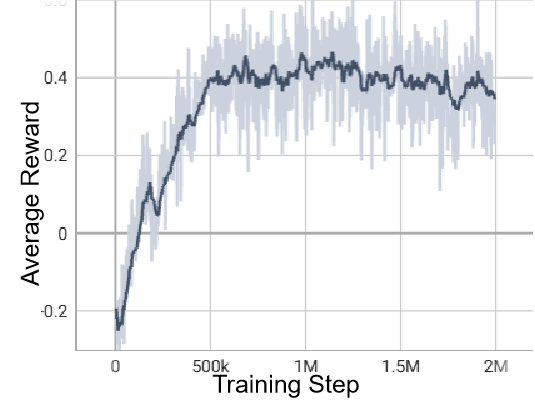}%
		\label{fig8_case2}}
	\caption{(a) Pursuit UAV pursuit-evasion game training reward curve. (b) Bait UAV pursuit-evasion game training reward curve.}
	\label{fig_8}
\end{figure}

Additionally, Fig. 9 presents a comparison of MEADDQN with PER to other reinforcement learning algorithms using the training of the pursuit UAV as an example.

\begin{figure}[!t]
	\centering
	\includegraphics[width=3.3in]{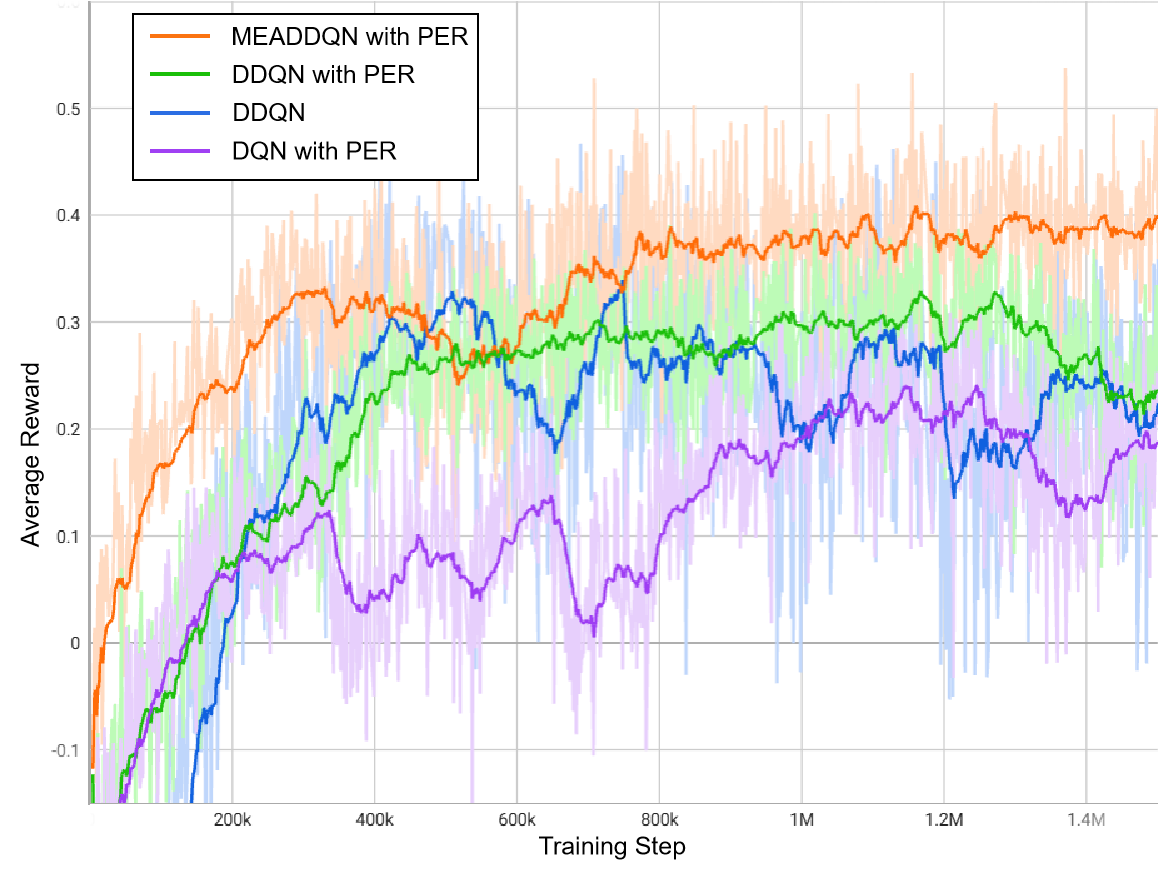}
	\caption{Comparison of reinforcement learning algorithms in pursuit-evasion game.}
	\label{fig_9}
\end{figure}

Fig. 10 demonstrates the track simulation results of the pursuit-evasion game of two different roles against matrix game algorithm, and gives real-time reward curves. As Fig. 10(a) shows, the Pursuit UAV can perform the pursuit-evasion game task very well, consistently maintaining a dominant position throughout the pursuit process. In the dogfight where both sides gain similar rewards, the pursuit UAV can adjust to secure the dominant position, thereby widening the reward gap and ultimately intercepting the blue UAV. In the test, the interception reward is set to $r_{final}=2$. The target of the bait UAV consistently receives positive reward feedback in Fig. 10(b), because the bait UAV in the trajectory simulation always stays in a position that appears to be advantageous to its target, Moreover, the special reward calculation method for the bait UAV enables it to receive a high reward under these circumstances.

\begin{figure*}[!b]
	\centering
	\subfloat[]{\includegraphics[width=3.5in]{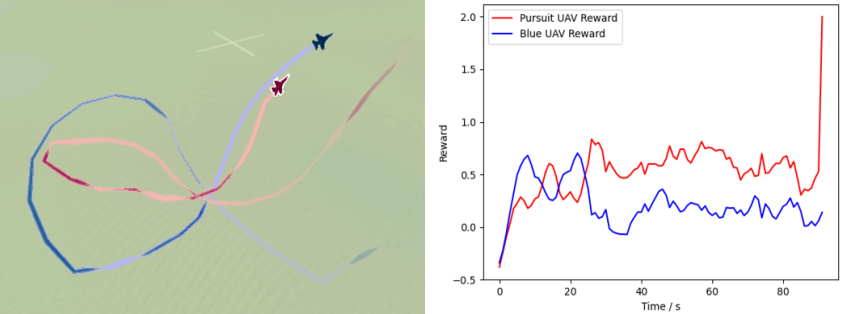}%
		\label{fig10_case1}}
	\hfil
	\subfloat[]{\includegraphics[width=3.5in]{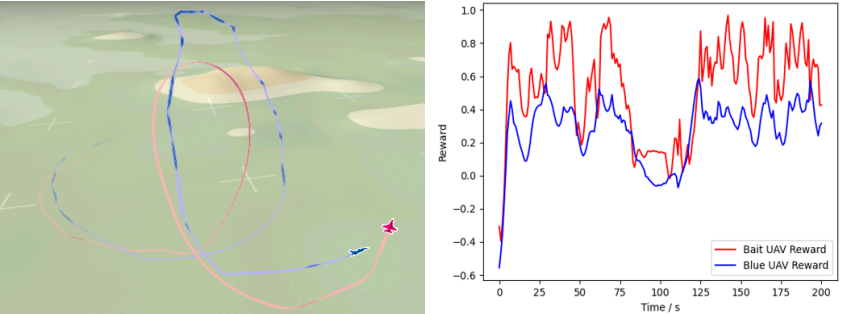}%
		\label{fig10_case2}}
	\caption{UAV pursuit-evasion game simulation tracks and real-time reward curves. (a) Pursuit UAV simulation results. (b) Bait UAV simulation results}
	\label{fig_10}
\end{figure*}

\subsection{Multi-Role UAV Cooperative Pursuit-Evasion Game.}

\begin{table}[!t]
	\caption{Role and Target Allocation Method\label{tab:table1}}
	\centering
	\begin{tabular}{|c|p{2.5in}|}
		\hline
		2v1 & The Red Team has a numerical advantage and does not need to worry about encirclement, so set both UAVs to  pursuit mode. \\ 
		\hline
		2v2 & Periodically, for each blue UAV, calculate the reward score for all red UAVs acting as pursuit UAVs in the current scenario, taking the higher group as the pursuit group and the other group as the bait group. Once the pursuit group completes its task, switch to a 2v1 mode. \\ 
		\hline
		3v2 & Periodically, for each blue UAV, calculate the reward score for all red UAVs acting as pursuit UAVs in the current scenario, taking the highest 2v1 group as the pursuit group and the other group as the bait group. Once the pursuit group completes its task, the scenario transitions to a 3v1 mode with three pursuit UAVs. \\ 
		\hline
	\end{tabular}
\end{table}

\begin{figure*}[!t]
	\centering
	\includegraphics[width=6.7in]{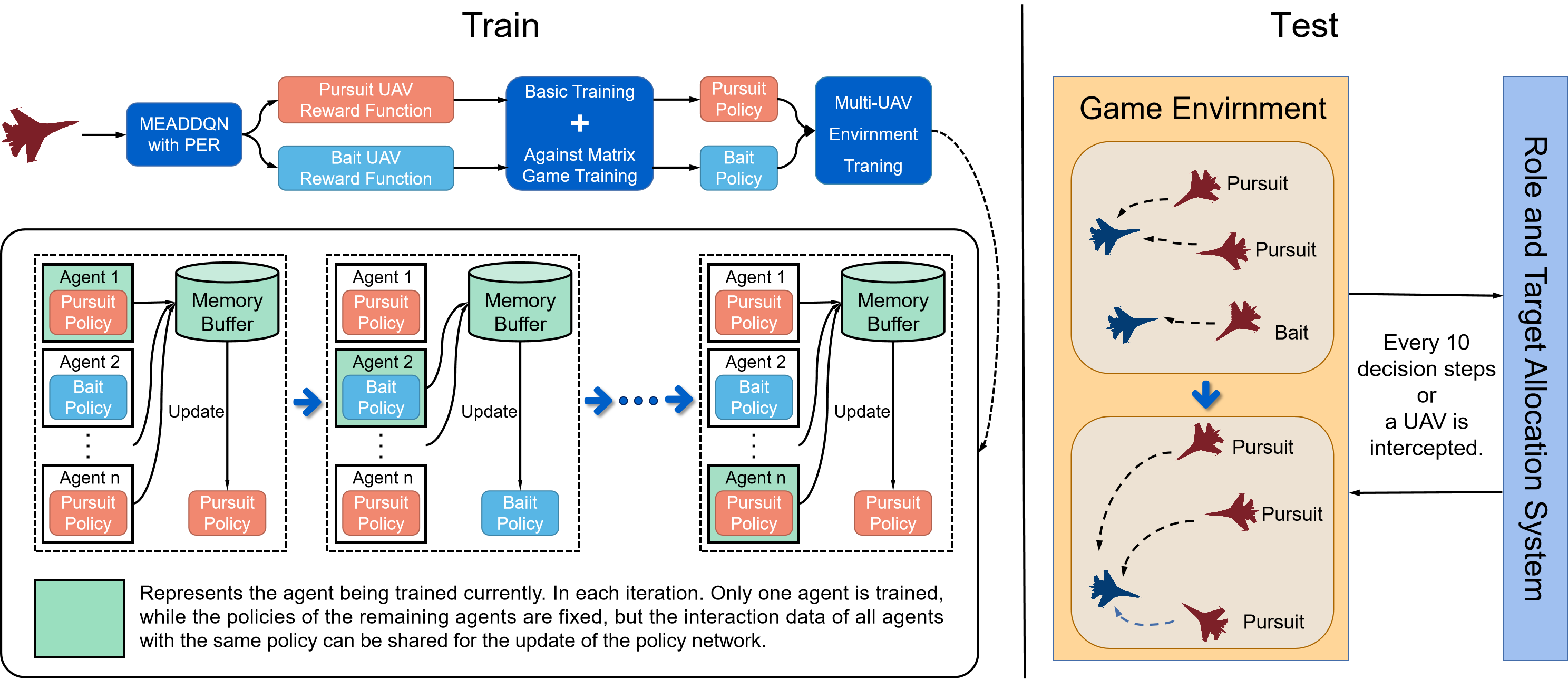}
	\caption{Multi-role UAV cooperative pursuit-evasion game framework.}
	\label{fig_11}
\end{figure*}

\begin{figure*}[!t]
	\centering
	\subfloat[]{\includegraphics[width=2.1in]{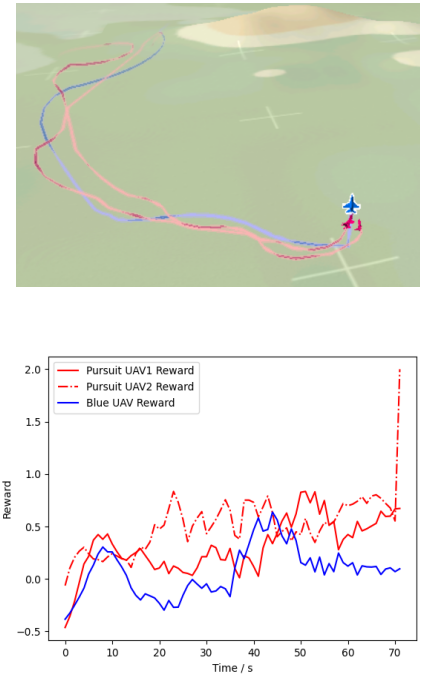}%
		\label{fig12_case1}}
	\hfil
	\subfloat[]{\includegraphics[width=2.1in]{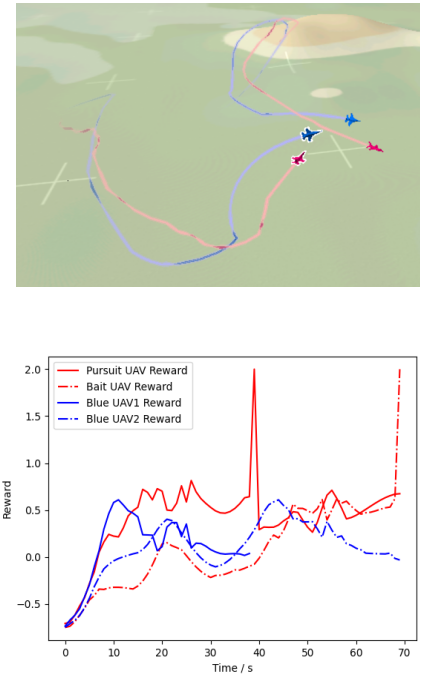}%
		\label{fig12_case2}}
	\hfil
	\subfloat[]{\includegraphics[width=2.1in]{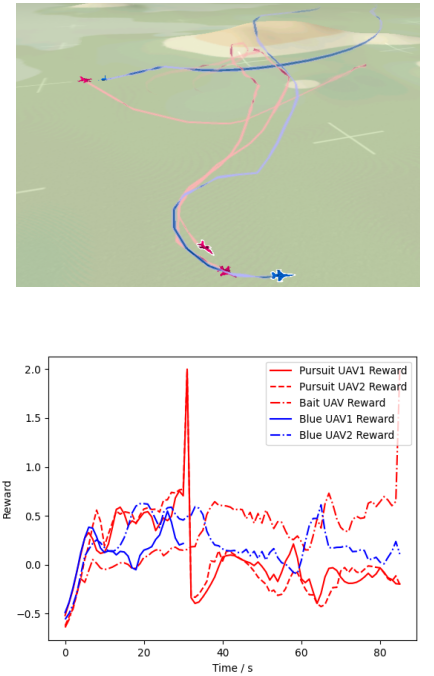}%
		\label{fig12_case3}}
	\caption{Multi-UAV cooperative pursuit-evasion game simulation tracks and real-time reward curves. (a) 2v1 simulation results. (b) 2v2 simulation results. (c) 3v2 simulation results.}
	\label{fig_12}
\end{figure*}

After training pursuit UAV and bait UAV, two different roles of UAV, this study proposed a Multi-role UAV cooperative pursuit-escape game method, which also uses matrix game algorithm as the opponent for pursuit-escape game. According to the Multi-UAV pursuit-escape game environment introduced in Section $\text{II}$, in the Multi-role UAV cooperative pursuit-escape game, the red UAVs are initialized with different roles and execute different strategies, while the blue UAVs are controlled by matrix game algorithm. Once the red UAV has determined its own role, it must also establish a clear target, whether it be pursuit or bait. The roles and targets of each UAV can be assigned based on the current game situation. In order to enhance the efficiency of the game, this study prioritizes allocation methods that can quickly achieve tail-biting scenarios, enabling pursuit UAV to quickly eliminate Blue Team members. Bait UAV are responsible for creating one-on-one or multi-on-one scenarios to enable pursuit UAV to avoid potential encirclement. 

Fig. 11 shows the training and testing procedure of multi-role UAV cooperative pursuit-evasion game. In training procedure, the pursuit and bait policies, which have been trained in the 1v1 UAV game, are concurrently applied in the multi-UAV environment and iterative trained through independent reinforcement learning. In each iteration, only one agent in the environment is designated as the training status, while the policies of the remaining agents are held constant. This approach effectively mitigates environmental fluctuations and enhances training stability. And as MEADDQN is an offline reinforcement learning algorithm, the interaction data from agents using the same policy as the agent being trained can also be utilized for training and updating the policy network, thereby enhancing sample efficiency. During the training process, only the UAVs’ target are allocated, while their roles remain unchanged after initialization. In the test, the roles and targets of the UAVs are allocated by the role and target allocation system every 10 decision steps or after a UAV is intercepted. The study conducted experiments in 2v1, 2v2, and 3v2 settings, implementing corresponding methods for role and target allocation as presented in Table $\text{I}$. the simulation results are illustrated in Fig. 12.

Fig. 12(a) shows that two red UAVs play the role of pursuit UAV, successfully tracking and interception the target about 70 seconds after the 2v1 game begin. Fig. 12(b) depicts a game scenario involving two red UAVs, wherein one is designated as the pursuit UAV and the other as the bait UAV. The bait UAV sacrifices itself to strategically create a favorable situation for the pursuit UAV.(The reward curves of simulation tests in Fig. 12 use the reward calculation methodology of the pursuit UAV to show the UAVs' advantage and disadvantage in a pursuit-evasion game.) Fig. 12(c) illustrates that at the beginning of the 3v2 game, two red UAVs function as pursuit UAVs while one red UAV serves as a bait UAV. The target has been successfully intercepted by the pursuit group approximately 30 seconds into the game. At this moment, the bait UAV transitions into pursuit mode and efficiently accomplishes the interception of the target. Table $\text{II}$, $\text{III}$, and $\text{IV}$ shows the winning rates of multi-UAV pursuit-evasion game over 100 test episodes with each episode lasting 1 minute, 3 minutes, and 5 minutes.

\begin{table}[!t]
	\caption{Win Rate of Pursuit-Evasion Game in 100 Test Episodes (1 Min)\label{tab:table1}}
	\centering
	\begin{tabular}{|c|c|c|c|}
            \hline
		\   &  \ \ win \ \ \ & \ \ standoff \ \ & \ \  lose \ \ \\
		\hline
		  \ \ 2v1 \ \ &  26\%  &  74\%  & 0\% \\ 
		\hline
		  \ \ 2v2 \ \ &  11\%  &  89\%  & 0\% \\ 
            \hline
		  \ \ 3v2 \ \ &  18\%  &  82\%  & 0\% \\ 
            \hline
	\end{tabular}
\end{table}

\begin{table}[!t]
	\caption{Win Rate of Pursuit-Evasion Game in 100 Test Episodes (3 Min)\label{tab:table1}}
	\centering
	\begin{tabular}{|c|c|c|c|}
            \hline
		\   &  \ \ win \ \ \ & \ \ standoff \ \ & \ \  lose \ \ \\
		\hline
		\ \ 2v1 \ \ &  54\%  &  46\%  & 0\% \\ 
		\hline
		  \ \ 2v2 \ \ &  27\%  &  73\%  & 0\% \\ 
            \hline
		  \ \ 3v2 \ \ &  35\%  &  65\%  & 0\% \\ 
            \hline
	\end{tabular}
\end{table}

\begin{table}[!t]
	\caption{Win Rate of Pursuit-Evasion Game in 100 Test Episodes (5 Min)\label{tab:table1}}
	\centering
	\begin{tabular}{|c|c|c|c|}
            \hline
		\   &  \ \  win \ \  & \ \  standoff \ \  & \ \  lose \ \  \\
		\hline
		  \ \  2v1 \ \  &  89\%  &  11\%  & 0\% \\ 
		\hline
		  \ \  2v2 \ \  &  65\%  &  35\%  & 0\% \\ 
            \hline
		  \ \  3v2 \ \  &  76\%  &  24\%  & 0\% \\
            \hline
	\end{tabular}
\end{table}

\section{Conclusion}
This paper proposes the MEADDQN with PER algorithm to address the problem of multi-UAV pursuit-evasion game. By assigning distinct tasks to UAVs in the pursuit-evasion game environment, We trained two types of UAVs with different policies, enabling them to collaboratively solve the multi-drone pursuit-evasion game problem through collaboration and task allocation. The method proposed in this paper enhances the mission efficiency and cooperation ability of multi-UAV pursuit-evasion game. The future will witness further exploration of multi-UAV cooperation scenarios, with the aim of proposing universally applicable methodology for role and target allocation. In addition, the challenge of autonomous decision-making in the presence of incomplete information, particularly when the UAV lacks a comprehensive perception or encounters communication obstacles, constitutes a primary focus for our forthcoming studies.


\end{document}